\documentclass[11pt,a4paper]{article}
\usepackage[hyperref]{emnlp2018}
\usepackage{times}
\usepackage{latexsym}

\usepackage{url}

\usepackage{booktabs}
\usepackage{graphicx}
\usepackage{amsmath}
\usepackage{amssymb}
\usepackage{multirow}

\aclfinalcopy 

\title{Learning Universal Sentence Representations \\with Mean-Max Attention Autoencoder}

\author{Minghua Zhang, Yunfang Wu\thanks{Corresponding author.}, Weikang Li \and Wei Li \\
Key Laboratory of Computational Linguistics, Ministry of Education \\
School of Electronics Engineering and Computer Science, Peking University, Beijing, China \\
{\tt \{zhangmh,wuyf,wavejkd,liweitj47\}@pku.edu.cn}
}

\date{}

\begin{document}
\maketitle
\begin{abstract}
In order to learn universal sentence representations, previous methods focus on complex recurrent neural networks or supervised learning. In this paper, we propose a mean-max attention autoencoder (mean-max AAE) within the encoder-decoder framework. Our autoencoder rely entirely on the MultiHead self-attention mechanism to reconstruct the input sequence. In the encoding we propose a mean-max strategy that applies both mean and max pooling operations over the hidden vectors to capture diverse information of the input. To enable the information to steer the reconstruction process dynamically, the decoder performs attention over the mean-max representation. By training our model on a large collection of unlabelled data, we obtain high-quality representations of sentences. Experimental results on a broad range of 10 transfer tasks demonstrate that our model outperforms the state-of-the-art unsupervised single methods, including the classical skip-thoughts \cite{kiros2015skip} and the advanced skip-thoughts+LN model \cite{ba2016layer}. Furthermore, compared with the traditional recurrent neural network, our mean-max AAE greatly reduce the training time. \footnote{Our code is publicly available at \url{https://github.com/Zminghua/SentEncoding}.}
\end{abstract}

\section{Introduction}
\label{sec:introduction}

To automatically get the distributed representations of texts (words, phrases and sentences) is a fundamental task for natural language processing (NLP). There have been efficient learning algorithms to acquire the representations of words \cite{mikolov2013efficient}, which have shown to provide useful features for various tasks. Interestingly, the acquired word representations reflect some observed aspects of human conceptual organization \cite{hill2015simlex}. In recent years, learning sentence representations has attracted much attention, which is to encode sentences into fixed-length vectors that could capture the semantic and syntactic properties of sentences and can then be transferred to a variety of other NLP tasks.

The most widely used method is to employ an encoder-decoder architecture with recurrent neural networks (RNN) to predict the original input sentence or surrounding sentences given an input sentence \cite{kiros2015skip,ba2016layer,hill2016learning,gan2017learning}. However, the RNN becomes time consuming when the sequence is long. The problem becomes more serious when learning general sentence representations that needs training on a large amount of data. For example, it took two weeks to train the skip-thought \cite{kiros2015skip}. Moreover, the traditional RNN autoencoder generates words in sequence conditioning on the previous ground-truth words, i.e., teacher forcing training \cite{williams1989learning}. This teacher forcing strategy has been proven important because it forces the output of the RNN to stay close to the ground-truth sequence. However, at each time step, allowing the decoder solely to access the previous ground-truth words weakens the encoder's ability to learn the global information of the input sequence.

Some other approaches \cite{conneau2017supervised,cer2018universal,subramanian2018learning} attempt to use the labelled data to build a generic sentence encoder, such as the Stanford Natural Language Inference (SNLI) dataset \cite{bowman2015large}, but such large-scale high-quality labelled data appropriate for training sentence representations is generally not available in other languages.

In this paper, we are interested in learning universal sentence representations based on a large amount of naturally occurring corpus, without using any labelled data. We propose a mean-max attention autoencoder (mean-max AAE) to model sentence representations. Specifically, an encoder performs the MultiHead self-attention on an input sentence, and then the combined mean-max pooling operation is employed to produce the latent representation of the sentence. The representation is then fed into a decoder to reconstruct the input sequence, which also depends entirely on the MultiHead self-attention. At each time step, the decoder performs attention operations over the mean-max encoding, which on the one hand, enables the decoder to utilize the global information of the input sequence rather than generating words solely conditioning on the previous ground-truth words, and on the other hand, allows the decoder to attend to different representation subspaces dynamically.

We train our autoencoder on a large collection of unlabelled data, and evaluate the sentence embeddings across a diverse set of 10 transfer tasks. The experimental results show that our model outperforms the state-of-the-art unsupervised single models, and obtains comparable results with the combined models. Our mean-max representations yield consistent performance gain over the individual mean and max representations. At the same time, our model can be efficiently parallelized and so achieves significant improvement in computational efficiency.

In summary, our contributions are as follows:
\begin{itemize}
\item We apply the MultiHead self-attention mechanism to train autoencoder for learning universal sentence representations, which allows our model to do processing parallelization and thus greatly reduce the training time in large unlabelled data.
\item we adopt a mean-max representation strategy in the encoding and then the decoder conducts attention over the latent representations, which can well capture the global information of the input from different views.
\item After training only on naturally occurring unordered sentences, we obtain a simple and fast sentence encoder, which is an unsupervised single model and achieves state-of-the-art performance on various transfer tasks.
\end{itemize}

\section{Related Work}
\label{sec:relatedwork}

With the flourishing of deep learning in NLP research, a variety of approaches have been developed for mapping word embeddings to fixed-length sentence representations. The methods generally fall into the following categories.

{\bf Unsupervised training with unordered sentences.} This kind of methods depends only on naturally occurring individual sentences. \citet{le2014distributed} propose the paragraph vector model, which incorporates a global context vector into the log-linear neural language model \cite{mikolov2013distributed}, but at test time, inference needs to be performed to compute a new vector. \citet{arora2017simple} propose a simple but effective Smooth Inverse Frequency (SIF) method, which represents sentence by a weighted average of word embeddings. \citet{hill2016learning} introduce sequential denoising autoencoders (SDAE), which employ the denoising objective to predict the original source sentence given a corrupted version. They also implement bag-of-words models such as word2vec-SkipGram, word2vec-CBOW. Our model belongs to this group, which has no restriction on the required training data and can be trained on sets of sentences in arbitrary order.

{\bf Unsupervised training with ordered sentences.} This kind of method is trained to predict the surrounding sentences of an input sentence, based on the naturally occurring coherent texts. \citet{kiros2015skip} propose the skip-thoughts model, which uses an encoder RNN to encode a sentence and two decoder RNN to predict the surrounding sentences. The skip-thought vectors perform well on several tasks, but training this model is very slow, requiring several days to produce meaningful results. \citet{ba2016layer} further obtain better results by adding layer-norm regularization on the skip-thoughts model. \citet{gan2017learning} explore a hierarchical model to predict multiple future sentences, using a convolutional neural network (CNN) encoder and a long-short term memory (LSTM) decoder. \citet{logeswaran2018efficient} reformulate the problem of predicting the context in which a sentence appears as a classification task. Given a sentence and its context, a classifier distinguishes context sentences from other contrastive sentences based on their vector representations.

{\bf Supervised learning of sentence representations.} \citet{hill2016learning} implement models trained on supervised data, including dictionary definitions, image captions from the COCO dataset \cite{Lin2014Microsoft} and sentence-aligned translated texts. \citet{conneau2017supervised} attempt to exploit the SNLI dataset for building generic sentence encoders. Through examining $7$ different model schemes, they show that a bi-directional LSTM network with the max pooling yields excellent performance. \citet{cer2018universal} apply multi-task learning to train sentence encoders, including a skip-thought like task, a conversational input-response task and classification tasks from the SNLI dataset. They also explore combining the sentence and word level transfer models. \citet{subramanian2018learning} also present a multi-task learning framework for sentence representations, and train their model on several data resources with multiple training objectives on over $100$ million sentences.

\section{Model Description}
\label{sec:modeldescription}

\begin{figure}
\centering
\includegraphics[scale=1.4]{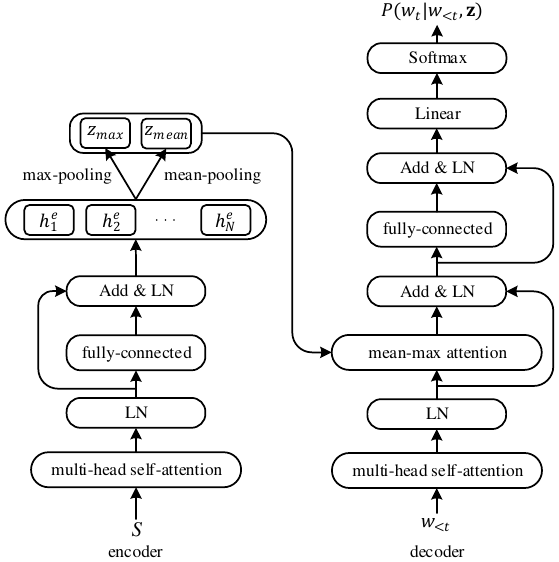}
\caption{\label{autoencoder} Illustration of the mean-max attention autoencoder.}
\end{figure}

Our model follows the encoder-decoder architecture, as shown in Figure~\ref{autoencoder}. The input sequence is compressed into a latent mean-max representation via an encoder network, which is then used to reconstruct the input via a decoder network.

\subsection{Notation}
\label{ssec:notation}

In our model, we treat the input sentence as one sequence of tokens. Let $S$ denote the input, which is comprised of a sequence of tokens $\{w_1, w_2, \dots, w_N\}$, where $N$ denotes the length of the sequence. An additional ``$<$$/S$$>$'' token is appended to each sequence. Each word $w_t$ in $S$ is embedded into a $k$-dimensional vector $e_t=W_e[w_t]$, where $W_e \in \mathbb{R}^{d_w \times V}$ is a word embedding matrix, $w_t$ indexes one element in a $V$-dimensional set (vocabulary), and $W_e[v]$ denotes the $v$-th column of matrix $W_e$.

In order for the model to take account of the sequence order, we also add ``positional encodings'' \cite{Vaswani2017Attention} to the input embeddings:
\begin{align}
\label{equ psin}
& p_t[2i] = sin(\frac{t}{10000^{2i/d_w}}) \\
\label{equ pcos}
& p_t[2i+1] = cos(\frac{t}{10000^{2i/d_w}})
\end{align}
where $t$ is the position and $i$ is the dimension. Each dimension of the positional encoding corresponds to a sinusoid. Therefore, the input of our model can be represented as $x_t = e_t + p_t$.

In the following description, we use $h^e_t$ and $h^d_t$ to denote the hidden vectors of the encoder and decoder respectively, the subscripts of which indicate timestep $t$, and the superscripts indicate operations at the encoding or decoding stage.

\subsection{MultiHead Self-Attention}
\label{ssec:multiheadattention}

In this subsection, we give a quick overview of MultiHead Self-Attention mechanism \cite{Vaswani2017Attention}. The attention is to map a query $q$ and a set of key-value pairs $(K, V)$ to an output. The output is computed as a weighted sum of the values, where the weight assigned to each value is computed based on the query and the corresponding key. The MultiHead mechanism applies multiple attention operations in parallel. Given $q$ and $(K, V)$, we can obtain the attention vector $a$ by:
\begin{align}
& a = MultiHead(q, K, V)\label{eq:multi} \\
&{\ }{\;} = concat(head_1, \dots, head_l)\label{eq:concat} \\
& head_i = attention(\overline{q}, \overline{K}, \overline{V})\label{eq:att} \\
&{\qquad}{\ } = softmax( \frac{\overline{q} \overline{K}^T}{ \sqrt{d_k} } ) \overline{V} \label{eq:softmax}
\end{align}
where
\begin{align}
& \overline{q}, \overline{K}, \overline{V} = q W^q_i, K W^K_i, V W^V_i\label{eq:dot}
\end{align}
$W^q_i$, $W^K_i$ and $W^V_i$ are parameter matrices; $\overline{q} \in \mathbb{R}^{d_k}$, $\overline{K} \in \mathbb{R}^{n_k \times d_k}$ and $\overline{V} \in \mathbb{R}^{n_k \times d_v}$; $d_k$ and $d_v$ are the dimensions of $\overline{K}$ and $\overline{V}$ respectively; $n_k$ is the number of key-value pairs.

The MultiHead self-attention allows the model to jointly attend to information from different positions. Due to the reduced dimension of each head and parallel operations, the total computational cost is similar to that of a single-head attention.

\subsection{Attention Encoder}
\label{ssec:attentionencoder}

The encoder has two sub-layers. The first is a MultiHead self-attention mechanism, and the second is a position-wise fully connected feed-forward network which consists of two linear transformations with a ReLU activation in between. Different from \citet{Vaswani2017Attention}, we remove the residual connections in the MultiHead self-attention layer and only employ a residual connection in the fully connected layer, allowing the model to expand the dimension of hidden vectors to incorporate more information.

Given the input $\mathbf{x} = (x_1, \dots, x_N)$, the hidden vector $h^e_t$ at time-step $t$ is computed by:
\begin{align}
\label{equ ea}
& a^e_t = MultiHead(x_t, \mathbf{x}, \mathbf{x}) \\
\label{equ ealn}
& \overline{a}^e_t = LN(a^e_t) \\
\label{equ eh}
& \overline{h}^e_t = max(0, \overline{a}^e_t W^e_1 + b^e_1) W^e_2 + b^e_2 \\
\label{equ ehln}
& h^e_t = LN(\overline{h}^e_t + \overline{a}^e_t)
\end{align}
where $W^e_1 \in \mathbb{R}^{d_m \times d_f}$ and $W^e_2 \in \mathbb{R}^{d_f \times d_m}$ are parameter matrices; $b^e_1 \in \mathbb{R}^{d_f}$ and $b^e_2 \in \mathbb{R}^{d_m}$ are bias vectors; $d_m$ and $d_f$ are the dimensions of hidden vector and fully connected inner layer respectively; $LN$ denotes layer normalization.

Our model can be efficiently parallelized over the whole input. We can obtain all hidden vectors $(h^e_1, \dots, h^e_N)$ simultaneously for an input sequence, thus greatly reducing the computational complexity compared with the sequential processing of LSTM.

\subsection{Mean-Max Representation}
\label{ssec:meanmaxrepresentation}

Given the varying number of hidden vectors $\{h^e_t\}_{t = [1, \dots, N]}$, we need to transform these local hidden vectors into a global sentence representation. We would like to apply the pooling strategy, which makes the extracted representation independent of the length of the input sequence and obtains a fixed-length vector. \citet{conneau2017supervised} examine BiLSTM with mean and max pooling for fixed-size sentence representation, and they conclude that the max pooling operation performs better on transfer tasks.

In this work, we propose to apply mean and max pooling simultaneously. The max pooling takes the maximum value over the sequence, which tries to capture the most salient property while filtering out less informative local values. On the other hand, the mean pooling does not make sharp choices on which part of the sequence is more important than others, and so it captures general information while not focusing too much on specific features. Obviously, the two pooling strategies can complement each other. The mean-max representation is obtained by:
\begin{align}
\label{equ max}
& z_{max}[i] = \max_{t} h^e_{ti} \\
\label{equ mean}
& z_{mean} = \frac{1}{N}\sum_{t}{h^e_t} \\
\label{equ z}
& \mathbf{z} = [z_{max}, z_{mean}]
\end{align}

Through combining two different pooling strategies, our model enjoys the following advantages. First, in the encoder, we can summarize the hidden vectors from different perspectives and so capture more diverse features of the input sequence, which will bring robustness on different transfer tasks. Second, in the decoder (as described in the next subsection), we can perform attention over the mean-max representation rather than over the local hidden vectors step by step, which would potentially make the autoencoder objective trivial.

\subsection{Attention Decoder}
\label{ssec:attentiondecoder}

As with the encoder, the decoder also applies the MultiHead self-attention to reconstruct the input sequence. As shown in Figure~\ref{autoencoder}, the encoder and decoder are connected through a mean-max attention layer, which performs attention over the mean-max representation generated by the encoder.

To facilitate expansion of the hidden size, we employ residual connections in the mean-max attention layer and the fully connected layer, but not in the MultiHead self-attention layer. Given $\mathbf{y} = (x_1, \dots, x_{t-1})$ and $\mathbf{z}$ as the decoder input, the hidden vector $h^d_t$ at time step $t$ is obtained by:
\begin{align}
\label{equ da}
& a^d_t = MultiHead(y_t, \mathbf{y}, \mathbf{y}) \\
\label{equ daln}
& \overline{a}^d_t = LN(a^d_t) \\
\label{equ za}
& a^z_t = MultiHead(\overline{a}^d_t, \mathbf{z}, \mathbf{z}) \\
\label{equ zaln}
& \overline{a}^z_t = LN(a^z_t + \overline{a}^d_t) \\
\label{equ dh}
& \overline{h}^d_t = max(0, \overline{a}^z_t W^d_1 + b^d_1) W^d_2 + b^d_2 \\
\label{equ dhln}
& h^d_t = LN(\overline{h}^d_t + \overline{a}^z_t)
\end{align}
where $W^d_1 \in \mathbb{R}^{d_m \times d_f}$ and $W^d_2 \in \mathbb{R}^{d_f \times d_m}$ are parameter matrices; $b^d_1 \in \mathbb{R}^{d_f}$ and $b^d_2 \in \mathbb{R}^{d_m}$ are bias vectors. $\mathbf{z}$ in Equation \eqref{equ za} is the mean-max representation generated by Equation \eqref{equ z}.

Given the hidden vectors $(h^d_1, \dots, h^d_N)$, the probability of generating a sequence $S$ with length-$N$ is defined as:
\begin{align}
\label{equ pw}
& P(w_t|w_{<t}, \mathbf{z}) \propto exp(W^d_3 h^d_t + b^d_3) \\
\label{equ loss}
& J(\theta) = {\sum_{t}{log P(w_t|w_{<t}, \mathbf{z})}}
\end{align}
The model learns to reconstruct the input sequence by optimizing the objective in Equation \eqref{equ loss}.

\section{Evaluating Sentence Representations}
\label{sec:evaluatingsentencerepresentations}

In the previous work, researchers evaluated the distributed representations of sentences by adding them as features in transfer tasks \cite{kiros2015skip,gan2017learning,conneau2017supervised}. We use the same benchmarks and follow the same procedure to evaluate the capability of sentence embeddings produced by our generic encoder.

\subsection{Transfer Tasks}
\label{ssec:transfertasks}

We conduct extensive experiments on $10$ transfer tasks. We first study the classification task on $6$ benchmarks: movie review sentiment (MR, SST) \cite{Bo2005Seeing,socher2013recursive}, customer product reviews (CR) \cite{Hu2004Mining}, subjectivity/objectivity classification (SUBJ) \cite{Bo2004A}, opinion polarity (MPQA) \cite{Wiebe2005Annotating} and question type classification (TREC) \cite{Li2002Learning}.

We also consider paraphrase detection on the Microsoft Research Paraphrase Corpus (MRPC) \cite{Dolan2004Unsupervised}, where the evaluation metrics are accuracy and F1 score.

We then evaluate on the SICK dataset \cite{Marelli2014A} for both textual entailment (SICK-E) and semantic relatedness (SICK-R). The evaluation metric is Pearson correlation for SICK-R. We also evaluate on the SemEval task of STS14 \cite{Agirre2014SemEval}, where the evaluation metrics are Pearson and Spearman correlations.

The processing on each task is as follows: 1) Employ the pre-trained attention autoencoder to encode all sentences into the latent mean-max representations. 2) Using the representations as features, apply the open SentEval with a logistic regression classifier \cite{conneau2017supervised} to automatically evaluate on all the tasks. For a fair comparison of the plain sentence embeddings, we adopt all the default settings.

\subsection{Experiment Setup}
\label{ssec:experimentsetup}

\begin{table*}[!ht]
\centering
\small
\begin{tabular}{l@{\quad}|c@{\quad}c@{\quad}c@{\quad}c@{\quad}c@{\quad}c@{\quad}c@{\quad}c@{\quad}c@{\quad}c}
\toprule
Method & MR & CR & SUBJ & MPQA & SST & TREC & MRPC & SICK-E & SICK-R & STS14 \\
\midrule
\midrule
\multicolumn{11}{l}{\emph{Unsupervised training of single model}} \\
\midrule
ParagraphVec (DBOW) & 60.2 & 66.9 & 76.3 & 70.7 & - & 59.4 & 72.9/81.1 & - & - & .42/.43 \\
SDAE & 74.6 & 78.0 & 90.8 & 86.9 & - & 78.4 & 73.7/80.7 & - & - & .37/.38 \\
SIF (GloVe + WR) & - & - & - & - & 82.2 & - & - & \textbf{84.6} & - & \textbf{.69/ -} \\
word2vec BOW$^\dag$ & 77.7 & 79.8 & 90.9 & 88.3 & 79.7 & 83.6 & 72.5/81.4 & 78.7 & 0.803 & \textbf{.65/.64} \\
GloVe BOW$^\dag$ & \textbf{78.7} & 78.5 & 91.6 & 87.6 & 79.8 & 83.6 & 72.1/80.9 & 78.6 & 0.800 & .54/.56 \\
FastSent & 70.8 & 78.4 & 88.7 & 80.6 & - & 76.8 & 72.2/80.3 & - & - & .63/.64 \\
FastSent+AE & 71.8 & 76.7 & 88.8 & 81.5 & - & 80.4 & 71.2/79.1 & - & - & .62/.62 \\
uni-skip & 75.5 & 79.3 & 92.1 & 86.9 & - & \textbf{91.4} & 73.0/81.9 & - & - & - \\
bi-skip & 73.9 & 77.9 & 92.5 & 83.3 & - & 89.4 & 71.2/81.2 & - & - & - \\
hierarchical-CNN & 75.3 & 79.3 & 91.9 & 88.4 & - & 90.4 & 74.2/\textbf{82.7} & - & - & - \\
composite-CNN & 77.1 & 80.6 & 92.1 & 88.6 & - & \textbf{91.2} & \textbf{74.8}/82.2 & - & - & - \\
skip-thoughts+LN$^\dag$ & \textbf{79.4} & \textbf{83.1} & \textbf{93.7} & \textbf{89.3} & \textbf{82.9} & 88.4 & - & 79.5 & \textbf{0.858} & .44/.45 \\
\midrule
mean-max AAE & \textbf{78.7} & \textbf{82.3} & \textbf{93.4} & \textbf{88.8} & \textbf{83.8} & \textbf{91.4} & \textbf{75.5/82.6} & \textbf{83.5} & \textbf{0.854} & .58/.56 \\
\midrule
\midrule
\multicolumn{11}{l}{\emph{Unsupervised training of combined model}} \\
\midrule
combine-skip$^\dag$ & 76.5 & 80.1 & 93.6 & 87.1 & 82.0 & 92.2 & 73.0/82.0 & 82.3 & 0.858 & .29/.35 \\
combine-CNN & 77.7 & 82.0 & 93.6 & 89.3 & - & 92.6 & 76.4/83.7 & - & - & - \\
\midrule
\midrule
\multicolumn{11}{l}{\emph{Trained on supervised data}} \\
\midrule
BiLSTM-Max (on SST) & * & 83.7 & 90.2 & 89.5 & * & 86.0 & 72.7/80.9 & 83.1 & 0.863 & .55/.54 \\
BiLSTM-Max (on SNLI) & 79.9 & 84.6 & 92.1 & 89.8 & 83.3 & 88.7 & 75.1/82.3 & 86.3 & 0.885 & .68/.65 \\
BiLSTM-Max (on AllNLI) & 81.1 & 86.3 & 92.4 & 90.2 & 84.6 & 88.2 & 76.2/83.1 & 86.3 & 0.884 & .70/.67 \\
\bottomrule
\end{tabular}
\caption{\label{table-tasks} Performance of sentence representation models on $10$ transfer tasks. Top $2$ results of unsupervised single models are shown in bold. The results with $^\dag$ are extracted from \citet{conneau2017supervised}.}
\end{table*}

We train our model on the open Toronto Books Corpus \cite{zhu2015aligning}, which was also used to train the skip-thoughts \cite{kiros2015skip} and skip-thoughts+LN \cite{ba2016layer}. The Toronto Book Corpus consists of $70$ million sentences from over $7,000$ books, which is not biased towards any particular domain or application.

The dimensions of hidden vectors and fully connected inner layer are set to $2,048$ and $4,096$ respectively. Hence, our mean-max AAE represents sentences with $4,096$ dimensional vectors. We set $l = 8$ parallel attention heads according to the development data.

We use the Adam algorithm \cite{kingma2014adam} with learning rate $2 \times 10^{-4}$ for optimization. Gradient clipping is adopted by scaling gradients when the norm of the parameter vector exceeds a threshold of $5$. We perform dropout \cite{srivastava2014dropout} and set the dropout rate to $0.5$. Mini-batches of size $64$ are used. Our model learns until the reconstruction accuracy in the development data stops improving.

Our aim is to learn a generic sentence encoder that could encode a large number of words. There always are some words that haven't been seen during training, and so we use the publicly available GloVe vectors \footnote{\url{https://nlp.stanford.edu/projects/glove/}} to expand our encoder's vocabulary. We set the word vectors in our models as the corresponding word vectors in GloVe, and do not update the word embeddings during training. Thus, any word vectors from GloVe can be naturally used to encode sentences. Our models are trained with a vocabulary of $21,583$ top frequent words in the Toronto Book corpus. After vocabulary expansion, we can now successfully cover $2,196,017$ words.

All experiments are implemented in Tensorflow \cite{abadi2016tensorflow}, using a NVIDIA GeForce GTX 1080 GPU with 8GB memory.

\begin{table}[!ht]
\centering
\small
\resizebox{!}{2.9cm}{
\begin{tabular}{l@{\ }|c@{\quad}c@{\quad}c}
\toprule
\multirow{2}{*}{Method} & \multicolumn{3}{c}{Macro/Micro} \\
& 8\_tasks & 7\_tasks & 6\_tasks \\
\midrule
\midrule
uni-skip & - & - & 83.0/83.7 \\
bi-skip & - & - & 81.4/82.1 \\
hierarchical-CNN & - & - & 83.3/84.0 \\
composite-CNN & - & - & 84.1/84.8 \\
skip-thoughts+LN & - & 85.2/85.9 & - \\
\midrule
mean-max AAE & \textbf{84.7}/\textbf{85.6} & \textbf{86.0}/\textbf{86.0} & 85.0/\textbf{85.9} \\
\midrule
\midrule
combine-skip & 83.4/84.2 & - & - \\
combine-CNN & - & - & \textbf{85.3}/85.8 \\
\midrule
\midrule
BiLSTM-Max(SST) & - & - & - \\
BiLSTM-Max(SNLI) & 85.0/86.2 & - & - \\
BiLSTM-Max(AllNLI) & 85.7/86.9 & - & - \\
\bottomrule
\end{tabular}}
\caption{\label{table-average} The macro and micro average accuracy across different tasks. The bold is the highest score among all unsupervised models.}
\end{table}

\subsection{Evaluation Results}
\label{ssec:evaluationresults}

A summary of our experimental results on $10$ tasks is given in Table~\ref{table-tasks}, in which the evaluation metric of the first $8$ tasks is accuracy. To make a clear comparison of the overall performance, we compute the ``macro'' and ``micro'' average of accuracy in Table~\ref{table-average}, where the micro average is weighted by the number of test samples in each task. In the previous work, different approaches conduct experiments on different benchmarks. Therefore we report the average scores on $6$ tasks, $7$ tasks and $8$ tasks, respectively.

We divide related models into three groups. The first group contains unsupervised single models, including the Paragraph Vector model \cite{le2014distributed}, the SDAE method \cite{hill2016learning}, the SIF model \cite{arora2017simple}, the FastSent \cite{hill2016learning}, the skip-thoughts (uni-skip and bi-skip) \cite{kiros2015skip}, the CNN encoder (hierarchical-CNN and composite-CNN) \cite{gan2017learning} and skip-thoughts+LN \cite{ba2016layer}. Our mean-max attention autoencoder sits in this group. The second group consists of unsupervised combined models, including combine-skip \cite{kiros2015skip} and combine-CNN \cite{gan2017learning}. In the third group, we list the results from the work of \citet{conneau2017supervised} only for reference, since it is trained on labelled data.

\textbf{Comparison with skip-thoughts+LN.} The skip-thoughts+LN is the best model among the existing single models. Compared with the skip-thoughts+LN, our method obtains better results on $4$ datasets (SST, TREC, SICK-E, STS14) and comparable results on $3$ datastes (SUBJ, MPQA, SICK-R). Looking at the STS14 results, we observe that the cosine metrics in our representation space is much more semantically informative than in skip-thoughts+LN representation space (pearson score of $0.58$ compared to $0.44$). Considering the overall performance shown in Table~\ref{table-average}, our model obtains better results both in the macro and micro average accuracy across $7$ considered tasks. In view of the required training data, the skip-thoughts+LN needs coherent texts while our model needs only individual sentences. Moreover, we train our model in less than $5$ hours on a single GPU compared to the best skip-thoughts+LN network trained for a month.

\textbf{Unsupervised combined models.} The results of the individual models \cite{kiros2015skip,gan2017learning} are not promising. To get better performance, they train two separate models on the same corpus and then combine the latent representations together. As shown in Table~\ref{table-average}, our mean-max attention autoencoder outperforms the classical combine-skip model by $1.3$ points in the average performance across $8$ considered tasks. Specially, the pearson correlation of our model is $2$ times over the combine-skip model on the STS14 task. Looking at the overall performance of $6$ tasks, our model gets comparable results with the combine-CNN, which combines the hierarchical and composite approaches to exploit the intra-sentence and inter-sentence information. Obviously, our model is simple and fast to implement compared with the combined methods.

\textbf{Supervised representation training.} It is unfair to directly compare our totally unsupervised model with the supervised representation learning method. \citet{conneau2017supervised} train the BiLSTM-Max (on ALLNLI) on the high-quality natural language inference data. Our model even performs better than the BiLSTM-Max (on ALLNLI) on the SUBJ and TREC tasks. More importantly, our model can be easily adapted to other low-resource languages.

\subsection{Model Analysis}
\label{ssec:modelanalysis}

Our model contains three main modules: the mean-max attention layer, the combined pooling strategy and the encoder-decoder network. We make a further study on these components. The experimental results are shown in Table~\ref{table-component}.

In our model, the mean-max attention layer allows the decoder to pay attention to the encoding representation of the full sentence at each time step dynamically. To summarize the contribution of the mean-max attention layer, we compare with traditional baselines, including the sequential denoising autoencoder (SDAE) with LSTM networks \cite{hill2016learning} and the CNN-LSTM autoencoder \cite{gan2017learning}, both of which only use the encoding representation to set the initial state of the decoder and follow the teacher forcing strategy.

We employ both the mean and max pooling operations over the local hidden vectors to obtain sentence embeddings. To validate the effectiveness of our mean-max representations, we train two additional models: (i) an attention autoencoder only with max pooling (max AAE) and (ii) an attention autoencoder only with mean pooling (mean AAE). The dimension of hidden vectors is also set to $2,048$.

Our encoder-decoder network depends on the MultiHead self-attention mechanism to reconstruct the input sequence. To test the effect of the MultiHead self-attention mechanism, we replace it with RNN and implement a mean-max RNN autoencoder (mean-max RAE) training on the same Toronto Books Corpus. A bidirectional LSTM computes a set of hidden vectors on an input sentence, and then the mean and max pooling operations are employed to generate the latent mean-max representation. The representation is then fed to a LSTM decoder to reconstruct the input sequence through attention operation over the latent representation. The parameter configurations are consistent with our other models. Moreover, we also train two additional models with different pooling strategies: mean RAE and max RAE.

\begin{table*}[!ht]
\centering
\small
\begin{tabular}{r@{\ }|c@{\quad}c@{\quad}c@{\quad}c@{\quad}c@{\quad}c@{\quad}c@{\quad}c@{\quad}c@{\quad}c@{\quad}c@{\ }}
\toprule
Method & Macro & MR & CR & SUBJ & MPQA & SST & TREC & MRPC & SICK-E & SICK-R & STS14 \\
\midrule
mean AAE & 84.1 & 78.6 & 81.6 & 93.3 & 88.6 & 83.5 & 90.2 & 74.8/82.3 & 82.2 & 0.851 & .57/.55 \\
max AAE & 84.1 & 78.6 & 81.4 & 93.1 & 88.7 & 82.8 & 90.6 & 75.1/82.5 & 82.2 & 0.851 & .57/.55 \\
mean-max AAE & 84.7 & 78.7 & 82.3 & 93.4 & 88.8 & 83.8 & 91.4 & 75.5/82.6 & 83.5 & 0.854 & .58/.56 \\
\midrule
\midrule
mean RAE & 83.1 & 77.1 & 80.2 & 92.4 & 88.4 & 81.9 & 88.8 & 74.3/82.2 & 81.9 & 0.836 & .66/.64 \\
max RAE & 83.0 & 77.2 & 80.1 & 92.0 & 88.4 & 81.0 & 88.6 & 74.3/82.0 & 82.0 & 0.851 & .68/.66 \\
mean-max RAE & 83.6 & 77.4 & 80.1 & 92.2 & 88.5 & 82.3 & 90.6 & 74.6/82.0 & 83.2 & 0.853 & .68/.65 \\
\midrule
\midrule
LSTM SDAE & - & 74.6 & 78.0 & 90.8 & 86.9 & - & 78.4 & 73.7/80.7 & - & - & .37/.38 \\
CNN-LSTM AE & - & 75.5 & 79.0 & 92.0 & 88.0 & - & 89.8 & 73.6/82.1 & - & - & - \\
\bottomrule
\end{tabular}
\caption{\label{table-component} Performance of different pooling strategies and different encoder-decoder networks on 10 transfer tasks. Macro is the macro average over the first 8 tasks whose metric is accuracy.}
\end{table*}

\textbf{Analysis on the mean-max attention layer.} Our mean-max attention layer brings significant performance gain over the previous autoencoders. Compared the mean RAE with LSTM-SDAE, both of which use the RNN-RNN encoder-decoder network to reconstruct the input sequence, our mean RAE consistently obtains better performance than LSTM-SDAE across all considered tasks. In particular, it yields a performance gain of $10.4$ on the TREC dataset and $29$ on the STS14 dataset. Compared with another CNN-LSTM autoencoder, our mean RAE also gets better performance for all but one task. It demonstrates that the mean-max attention layer enables the decoder to attend to the global information of the input sequence, thus go beyond the ``teacher forcing training''.

\textbf{Analysis on the pooling strategy.} Considering the overall performance, our mean-max representations outperform the individual mean and max representations both in the attention and RNN networks. In our attention autoencoder, the macro average score of the mean-max AAE is more than $0.6$ over the individual pooling strategy. In the RNN autoencoder, the combined pooling strategy yields a performance gain of $0.5$ over the mean pooling and $0.6$ over the max pooling. The results indicate that our mean-max pooling captures more diverse information of the input sequence, which is robust and effective in dealing with various transfer tasks.

\textbf{Comparison with RNN-based autoencoder.} As shown in Table~\ref{table-component}, our MultiHead self-attention network obtains obvious improvement over the RNN network in different sets of pooling strategies, and it yields a performance gain of $1.1$ when applying the best combined mean-max pooling operation. The results demonstrate that the MultiHead self-attention mechanism enables the sentence representations to capture more useful information about the input sequence.

\textbf{Analysis on computational complexity.} A self-attention layer connects all positions with a constant number of sequentially executed operations, whereas a recurrent layer requires $O(n)$ sequential operations. Therefore, our model greatly reduces the computational complexity. Excluding the number of parameters used in the word embeddings, the skip-thought model \cite{kiros2015skip} contains 40 million parameters, while our mean-max AAE has approximately 39 million parameters. It took nearly $50.4$ and $25.4$ minutes to train the skip-thought model \cite{kiros2015skip} and the skip-thoughts+LN \cite{ba2016layer} per $1000$ mini-batches respectively. Both the skip-thought and skip-thought+LN are implemented in Theano. A recent implementation of the skip-thoughts model was released by Google \footnote{\url{https://github.com/tensorflow/models/tree/master/research/skip_thoughts}}, which took nearly $25.9$ minutes to train $1000$ mini-batches on a GTX 1080 GPU. In our experiment, it took $3.3$ minutes to train the mean-max AAE model every $1000$ mini-batches.

\subsection{Attention Visualization}
\label{ssec:attentionvisualization}

\begin{figure}
\centering
\includegraphics[scale=0.68]{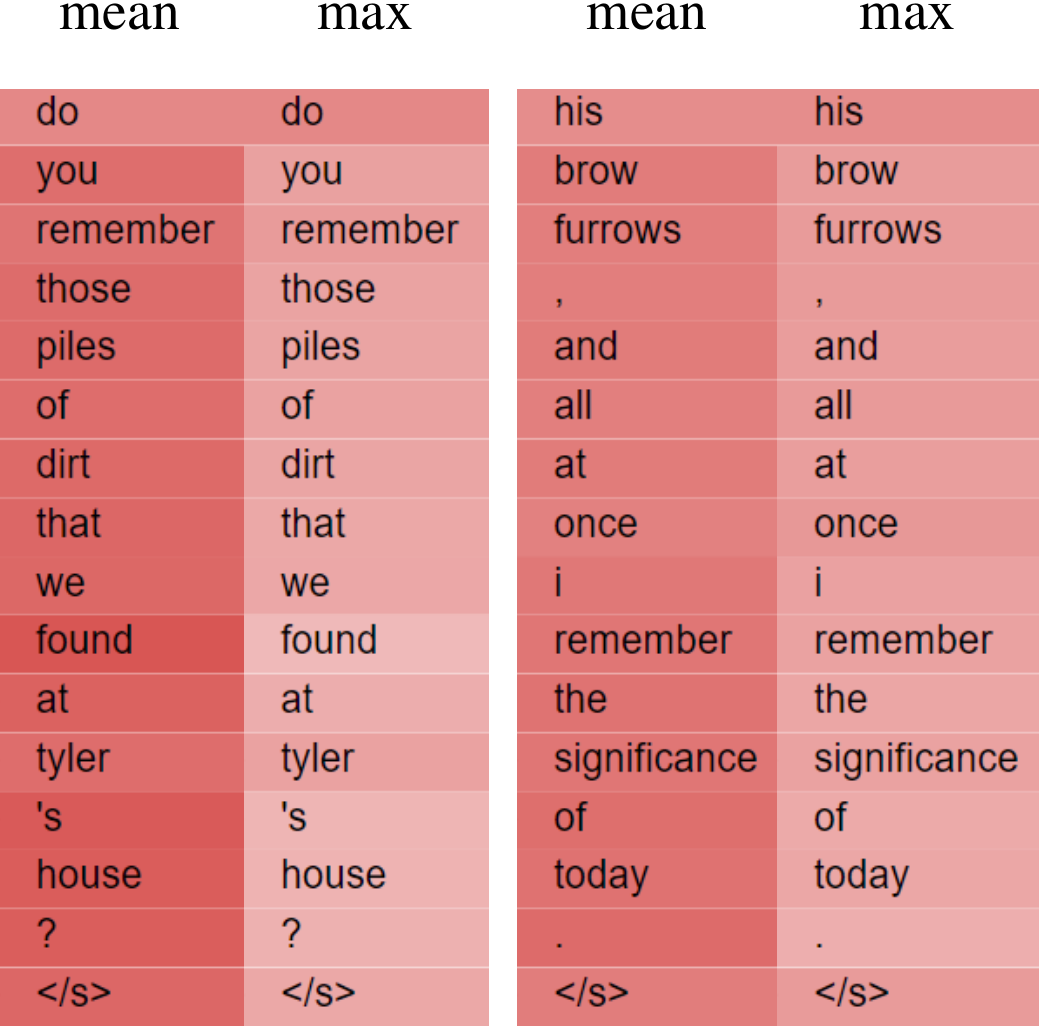}
\caption{\label{visualization} Two examples illustrate that our mean-max attention layer could attend to the two different representations dynamically.}
\end{figure}

The above experimental results have proven the effectiveness of the mean-max attention mechanism in the decoding. We further inspect the attention distributions captured by the mean-max attention layer, as shown in Figure~\ref{visualization}. The side-by-side heat illustrates how much the decoder pay attention to the mean representation and max representation respectively at each decoding step. We can see that the attention layer learns to selectively retrieve the mean or max representations dynamically, which relieve the decoder from the burden of generating words solely conditioning on the previous ground-truth words. Also, the two different representations can complement each other, and the mean representation plays a greater role.

\section{Conclusion}
\label{sec:conclusion}

In this paper, we present a mean-max AAE to learn universal sentence representations from unlabelled data. Our model applies the MultiHead self-attention mechanism both in the encoder and decoder, and employs a mean-max pooling strategy to capture more diverse information of the input. To avoid the impact of ``teacher forcing training'', our decoder performs attention over the encoding representations dynamically. To evaluate the effectiveness of sentence representations, we conduct extensive experiments on $10$ transfer tasks. The experimental results show that our model obtains state-of-the-art performance among the unsupervised single models. Furthermore, it is fast to train a high-quality generic encoder due to the paralleling operation. In the future, we will adapt our mean-max AAE to other low-resource languages for learning universal sentence representations.

\section*{Acknowledgments}
We would like to thank the anonymous reviewers for their valuable comments. This work is supported by the National Key Basic Research Program of China (2014CB340504) and National Natural Science Foundation of China (61773026, 61572245). The corresponding author of this paper is Yunfang Wu.

\bibliography{minghua}
\bibliographystyle{acl_natbib_nourl}

\end{document}